\documentclass{article}

\usepackage{PRIMEarxiv}

\usepackage[utf8]{inputenc} 
\usepackage[T1]{fontenc}    
\usepackage{hyperref}       
\usepackage{url}            
\usepackage{booktabs}       
\usepackage{amsfonts}       
\usepackage{nicefrac}       
\usepackage{microtype}      
\usepackage{lipsum}
\usepackage{fancyhdr}       
\usepackage{graphicx}       
\graphicspath{{media/}}     
\usepackage{amsmath}
\usepackage{subfig}
\usepackage{footnote}
\usepackage{tabularx}

\newcommand{\argmax}[1]{\underset{#1}{\operatorname{arg}\,\operatorname{max}}\;}

\title{Self-Attentive Constituency Parsing for UCCA-based Semantic Parsing}

\author{
  Necva Bölücü \\
  Computer Engineering Department \\
  Alparslan Türkeş Science and Technology University \\
  Adana\\
  \texttt{nbolucu@atu.edu.tr} \\
   \And
  Burcu Can \\
  Research Institute in Information and Language Processing \\
  University of Wolverhampton \\
  Wolverhampton\\
  \texttt{b.can@wlv.ac.uk} \\
}

\begin{document}
\maketitle

\begin{abstract}
Semantic parsing provides a way to extract the semantic structure of a text that could be understood by machines. It is utilized in various NLP applications that require text comprehension such as summarization and question answering. Graph-based representation is  one of the semantic representation approaches to express the semantic structure of a text. Such representations generate expressive and adequate graph-based target structures. 
    
In this paper, we focus primarily on UCCA graph-based semantic representation. The paper not only presents the existing approaches proposed for UCCA representation, but also proposes a novel self-attentive neural parsing model for the UCCA representation. We present the results for both single-lingual and cross-lingual tasks using zero-shot and few-shot learning for low-resource languages.
\end{abstract}


\section{Introduction}
Semantic parsing is a way to represent the semantic structure of a text. It aims to represent the meaning of a text in natural language that can be understood by machines. Parsing has long been dominated by tree-structured target representations. Graphs are receiving increasing attention in Natural Language Processing (NLP) in recent years due to their capability on expressing and generating adequate target structures for sentence-level grammatical analysis of the representation of the semantic structure of a text.  Following this, different schemes such as Abstract Meaning Representation (AMR; \cite{banarescu2013abstract}), bilexical Semantic Dependencies (SDP; \cite{oepen2016towards}), Universal Decompositional Semantics (UDS; \cite{white2016universal}), Parallel Meaning Bank (PMB; \cite{abzianidze2017parallel}), and finally Universal Conceptual Cognitive Annotation (UCCA; \cite{abend2013ucca}) have been proposed for semantic parsing. These graph-based representations, along with the advances on semantic parsing, have proven to be beneficial in many Natural Language Understanding (NLU) tasks, and have already demonstrated applicability to a variety of tasks, including summarization \cite{dohare2017text,liu2018toward}, paraphrase detection \cite{issa2018abstract}, machine translation \cite{song2019semantic}, question answering \cite{zeng2013using} and text simplification \cite{sulem2018simple}.

Over the past few years, NLP tasks have evolved in a different direction with the usage of transfer learning methods. Transfer learning aims to extract knowledge from a source setting and transfer it to a target setting. Cross-lingual learning and cross-framework learning are two paradigms that are used to transfer knowledge from one natural language or framework to another. The aim of knowledge transfer is to overcome the scarcity of data in a target language by compensating the deficiency with another language or dataset. With the cross-lingual representations of words, cross-lingual NLP tasks have emerged in machine translation \cite{kim2019effective} and semantic parsing \cite{zhang2017selective, zhang2018cross} in recent years. Cross-framework learning has gained attention for semantic parsing with an application on different frameworks \cite{li2019sjtu,oepen2019mrp, wang2020shanghaitech}. Existing studies on graph-based semantic parsing are either cross-lingual or cross-framework. This paper addresses the cross-lingual and cross-domain studies conducted only on the UCCA framework. First we discuss the studies conducted on UCCA-based semantic parsing, then we propose a neural model on UCCA representation.

The remaining of this paper is structured as follows: Section \ref{sec:background} describes the concept of semantic graph-based structure, Section \ref{sec:ucca} describes the details of the UCCA representation. We categorize and explain the approached applied to semantic parsing in Section \ref{sec:approaches} which is followed by the related work in Section \ref{sec:rel}. Section \ref{sec:shared} gives the details of the shared tasks conducted on graph-based frameworks including UCCA framework. Section \ref{sec:model} describes the proposed model, and Section \ref{sec:result} presents the experimental setting and results on the UCCA dataset, with an error analysis. Finally, Section \ref{sec:conc} concludes the paper with potential directions that remain as future work.

\section{Background}\label{sec:background}

Before describing the details of the UCCA framework, we provide background information on Semantic Graphs and Flavors from a general perspective.

\subsection{Semantic Graphs}

A semantic graph is a directed graph represented with a tuple $(T, N, E)$ where $N$ denotes a set of nodes,  $T \subset N$ is a set of root nodes and $E \subseteq N \times N$ denotes a set of edges. The number of coming and leaving edges from each node is represented as \textit{in} and \textit{out}-degree of a node, respectively. Semantic graphs can have multiple roots with an in-degree value of zero. 

Trees are connected (there exists an undirected path between any pair of nodes) and ought to be acyclic (contains no cycles). A tree consists of nodes with an in-degree value of zero apart from the root node. In contrast, semantic graphs can have nodes with an in-degree value of two or higher, and they can contain cycles (the directed path leading from a node to itself). Moreover, semantic graphs do not need to be connected.

\subsection{Flavors}

Semantic graph frameworks can be divided into three different types based on the relationships between the linguistic surface (string) and the nodes of the graph. This relationship is called anchoring or alignment \cite{koller2019graph}.

\begin{itemize}
    \item Flavor (0) is the strongest form of anchoring in which graph nodes injectively correspond to surface lexical units (i.e., tokens or words). In these graphs, each node is linked to one specific token and the nodes inherit the linear order of their corresponding tokens. EDS \cite{oepen2006discriminant}, SDP \cite{oepen2016towards} and DM \cite{ivanova2012did} are examples to Flavor (0) form.
    \item Flavor (1) is a more general form of the anchored semantic graphs. This type of semantic graphs is partially anchored (a subset of nodes are linked to the surface lexical units). This allows one-to-one correspondence to surface tokens or overlapping and sub-token or phrasal relationships with flexibility in the representation of meaning contributed by. UCCA \cite{abend2013ucca} is an example of this form.
    \item Flavor (2) is the weakest form of anchoring in which semantic graphs are simply unanchored. Semantic graphs do not consider the correspondence between nodes and the surface string as part of the representation of meaning. AMR \cite{banarescu2013abstract} is an example of this form.
\end{itemize}

\section{Universal Conceptual Cognitive Annotation (UCCA)}\label{sec:ucca}

UCCA is a novel cross-linguistically applicable  annotation scheme for encoding semantic annotations. It was developed by \cite{abend2013ucca}. This scheme is a multi-layer structure that allows extending it open-endedly. 

UCCA is represented by directed acyclic graphs (DAGs), whose leaves called terminals correspond to the tokens and multi-tokens in the text. The nodes of the graph are called units that can be either a terminal or several tokens that are jointly viewed as a single entity according to some semantic or cognitive consideration.  The edges of the graph point the role of the child in the relation (categories). 

UCCA is a multi-layered framework, where each layer corresponds to a “module” of semantic distinctions. The foundational layer of UCCA focuses on grammatically relevant information. It covers predicate-argument relations for predicates of all grammatical categories (verbal, nominal, adjectival and, others). The foundational layer views the text as a collection of \textit{Scenes}. A Scene can describe some movement or action, or otherwise a temporally persistent state. Scene contains relations that can be one or higher. The Relation can be \textit{Process} (P) if Scene evolves in time otherwise, it can be \textit{State} (S) that means Scene describes a temporally persistent state. \textit{Participants} (A) are the relations of the Scenes. They can be concrete or abstract. Embedded Scenes are also considered as Participant of the main Scenes. The secondary relations of the Scenes are marked as \textit{Adverbials} (D).

There are also non-Scene relations in the UCCA framework. \textit{Connectors} (N) are the relations between two or more entities with similar feature or types. The arguments of non-Scene relations are marked as \textit{Centers} (C). For example, in ``John and Mary", ``John" and ``Mary" are Centers and ``and" is Connector that connects two similar relations (Center). \textit{Elaborates} (E) is applied to a single argument. For example, in ``around a beam", `around" and ``a" are Elaborates that describe the ``beam".

The other type of relation between two or more units is \textit{Relator} (R) that does not evoke a Scene. In two different scenarios, the relation is marked as Relator. 1. A single entity is related to the other relations in the same context as Relator. For example ``in" in the phrase ``I saw cookies in the jar". Here ``in" relates ``the jar" to the rest of Scene. 2. Two units attached with different aspects of the same entity are related to Relator. For instance, ``of" relates ``suspicious" and ``people" in ``suspicious of people" that refers to the same entity.

There are also Function units in which terminals do not refer to a participant or relation. They function only as a part of the construction they are situated in. These are marked as \textit{Function} (F). For example, ``had" is a Function unit in Scene ``He had tied a ...". Linkage in UCCA refers to the relation between Scenes. In a sentence, there can be more than one Scene. There are three types of Linkage in the UCCA adopted from Basic Linguistic Theory \cite{dixon2010abasic}: 1. A Scene can be a Participant of the main Scene. In that case, it is marked as Participant (P) of the Scene. 2. A Scene can be an Elaborator of some unit in another Scene, in which it is marked as Elaborator (E). 3. There can be inter-Scene relations that are not covered above. The relation between these Scenes is marked as \textit{Linker} (L) and its arguments as \textit{Parallel Scenes} (H).

Moreover, some units express the speaker’s opinion of a Scene with specific words such as ``surprisingly", ``in my opinion". They constitute a Scene, and their participants are implicit. For such cases, a new category \textit{Ground} (G) is introduced.

\begin{figure}[t!]
\centering 
\resizebox*{13 cm}{!}{\includegraphics{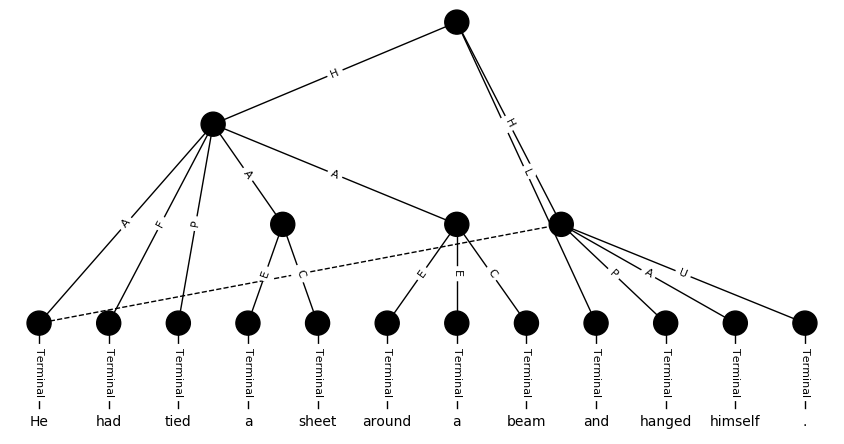}}
\caption{An Example of UCCA annotation for the sentence ``He has tied a sheet around a beam and hanged himself."}\label{fig:UCCA_annotation1}
\end{figure}

An example UCCA representation is illustrated in Figure \ref{fig:UCCA_annotation1}. In the example, there are two Scenes with a relation called \textbf{Process}: ``tied" and ``hanged". ``and" is a \textbf{Linker} between Scenes. \textbf{Participant} of the Process is the terminal ``He". ``He" has got two parents connected with one primary edge and one remote edge (dashed). The example also includes a discontinuous unit due to ``He" that is a Participant of the two Scenes.

The distinct features of the UCCA framework are be summarized as follows:
\begin{enumerate}
    \item There is a one-to-one correspondence between the terminal nodes and the spans in the sentence.
    \item One node may have multiple parents, among which one is annotated as the primary parent that is marked by solid line edges, and the others are annotated as remote parents and marked by dashed line edges. The primary edges form a tree structure, whereas the remote edges enable reentrancy, thereby forming a directed acyclic graph (DAG).
    \item UCCA involves nodes with discontinuous leaves, known as discontinuity. In the framework, some leaves are discontinuous because some terminal nodes it spans are not its descendants.
\end{enumerate}

\section{Parsing Approaches}\label{sec:approaches}
 
 Semantic parsing refers to the task of mapping natural language text to formal representations or abstractions of its meaning. We can categorize the methods applied to the semantic parsing task into four architectures: 

 \begin{enumerate}
     \item \textbf{Transition-Based Architectures:} In transition-based parsing, a semantic graph is generated with a series of actions similar to the transition-based dependency parsing. A graph $G = (V, E, l)$ is constructed by nodes and edges where $V$ is a set of nodes, $E$ is a set of edges, and $l : E \rightarrow L$ is a label function, with a set of labels $L$. At each step in parsing, the model predicts an action to take, which is possibly based on a  classifier. When parsing is complete, the sequence of parsing actions is used to deterministically reconstitute the meaning representation graph. This type of approach is mainly inspired by the transition-based dependency tree parsing \cite{nivre2008algorithms,yamada2003statistical}. To the best of our knowledge, \cite{sagae2008shift} are the first to apply this type of approach to predict predicate-argument structures grounded in Head-Driven Phrase Structure (HPSG) \cite{miyao2004corpus}. 
     
     \item \textbf{Factorization-Based Architectures:} A factorization-based parser explicitly models the target semantic structures by defining a score function that can evaluate the goodness of any candidate graph. The aim of this architecture to search for a subset $E' \subseteq E$ from a given a directed graph $G = (E,V)$ that represents an input sentence $x = w_1, \cdots, w_n$. The graph with the highest score is selected as follows: 
  \begin{equation}
      (V,E') = \argmax{G^* = (V, E^* \subseteq E)} SCORE_G(G^*)
  \end{equation}
  where $V$ is a set of surface tokens and $G$ is the corresponding graph.
  
  The set of possible graphs that can be used to generate an input sentence is usually large. Therefore, a parser needs to be informed to find the highest-scoring graph from the possible set of graphs. Generally, \textbf{factorization} strategy, defining a decomposable score function that enumerates all sub-parts of a candidate graph, is employed to solve the problem in the form of a combinatorial optimization. 
  
  To the best of our knowledge, the first proposed graph-based syntactic dependency parsing algorithm is proposed by \cite{mcdonald2006online}. \cite{kuhlmann2015parsing} generalize the graph-based parsing by Maximum Subgraph parsing with an application of Minimum Spanning Tree parsing.
  
  \item \textbf{Composition-Based Architectures:} Following the principle of compositionality, a semantic graph can be viewed as a derivation process where a set of lexical and syntactic-semantic rules is iteratively applied and evaluated.  From a linguistic perspective, such rules extensively encode explicit knowledge about natural languages. From a compositional perspective, such rules must be governed by a well-defined grammar formalism. 
   
\item \textbf{Translation-Based Architectures:} This type of approach is inspired by the success of neural sequence-to-sequence (seq2seq) models \cite{sutskever2014sequence} that form the basis of the Neural Machine Translation (NMT) systems. A translation-based parser encodes a semantic graph, and  relations of each word are viewed as a string in another language  \cite{buys2017robust,konstas2017neural,peng2017addressing}. 

 \end{enumerate}

\section{Related Work}\label{sec:rel}

UCCA framework \cite{abend2013ucca} is one of the recently proposed semantic graph representation frameworks. TUPA parser \cite{hershcovich2017transition} is the first parser proposed for generating UCCA representations. It is a neural transition-based parser model. The model involves various transition actions and features to handle discontinuous and remote nodes of UCCA graphs. \cite{hershcovich2018multitask} extend the TUPA parser with multi-task learning by utilizing other semantic graph representations such as AMR \cite{banarescu2013abstract}, UD \cite{nivre2017universal,nivre2016universal}, SDP \cite{oepen2016towards}.

UCCA framework was employed in several shared tasks. The first shared task is ``Cross-lingual Semantic Parsing with UCCA"  SemEval 2019  \cite{hershcovich2019semeval}. The shared task's goal is to leverage the semantic parsing in low-resource languages by using data in source-rich languages, thereby extending it as a cross-lingual parsing. The methods proposed within the shared task vary in terms of the parsing approaches. Also, some of the proposed methods handle remote edges by additional features, layers, or models.  T\"{u}Pa \cite{putz2019tupa} and CUNY-PekingU \cite{lyu2019cuny} adapt the baseline TUPA parser. T\"{u}Pa extends the feature set defined in the TUPA parser. It uses a feed-forward neural network along with ELMo contextualized embeddings. It handles remote edges by creating a set of features such as the last parser actions, named-entities, parts-of-speech and dependency types. CUNY-PekingU \cite{lyu2019cuny} introduces a Cascaded BiLSTM model that is also based on a TUPA parser with an extra MultiLayer Perceptron (MLP). CUNY-PekingU handles remote edges by adding a new node with a special type and removing them.

HLT@SUDA \cite{jiang2019hlt} converts UCCA graphs into constituency trees and applies span-based constituency parser as a parsing method, and uses a multi-task framework for learning remote edges.  It obtains the first place in English and German and the second place in the French. 

 DANGNT@UIT.VNU-HCM \cite{nguyen2019dangnt} is a rule-based system defined on the Stanford Parser. MaskParse@Deski\~n \cite{marzinotto2019maskparse} implements a recursive method with a bidirectional GRU and masking mechanism. UC-Davis \cite{yu2019uc} is an encoder-decoder model with an extra multi-layer introduced for remote edges. GCN-Sem \cite{taslimipoor2019gcn} proposes a sequence-to-sequence architecture composed of Graph Convolution Networks and BiLSTM components. Remote edges are ignored in their method due to the frequency of them compared the primary edges.

Additionally, the UCCA framework took place in Meaning Representation Parsing (MRP) cross-framework shared task \cite{oepen2020mrp,oepen2019mrp} in 2019 and 2020. While transition based approaches were widely used in MRP 2019, encoder-decoder was the most applied method in MRP 2020. Transition-based approaches in MRP 2019 \cite{bai2019sjtu,che2019hit,lai2019cuhk,straka2019ufal} tackled the problem with variations in the parsing process. SJTU \cite{bai2019sjtu} extends the TUPA parser by integrating all the frameworks into it by using shared features. HIT-SCIR \cite{che2019hit} extends the general transition-based parser by introducing a stack-LSTM to efficiently compute homogeneous operations within a batch. CUHK \cite{lai2019cuhk} includes a new action \textit{RESOLVE} that is used in all semantic frameworks (AMR, DM, PSD, EDC, UCCA),  \'UFAL MRPipe \cite{straka2019ufal} introduces a graph-to-graph architecture where the graph is built iteratively by two-layer wise transformations: adding nodes, and adding edges.

Graph-based approaches \cite{cao2019amazon,droganova2019ufal,koreeda2019hitachi,li2019sjtu,na2019jbnu,wang2019second,zhang2019suda} generally tackles the task as a search problem to find the graph with the highest score among all possible graphs for a given input. 
\cite{koreeda2019hitachi,li2019sjtu, na2019jbnu,zhang2019suda} utilize an LSTM encoder with a biaffine classifier to predict the edges between nodes. Amazon \cite{cao2019amazon} classifies meaning representations into two groups: lexical-anchoring and phrasal-anchoring. For UCCA, the authors apply a span-based CYK parsing by converting a graph into a constituent tree structure and use a unified latent-alignment-based parsing framework for DM, PSD, and AMR. \'UFAL-Oslo (\cite{droganova2019ufal}) applies third-party parsers (developed by other studies) instead of implementing a new parser. NeurboParser \cite{peng2017deep}, JAMR \cite{flanigan2016cmu,flanigan2014discriminative} and UDPipe \cite{straka2017tokenizing}, ShanghaiTech \cite{wang2019second} adopt a pointer-generator network \cite{zhang2019amr} in all frameworks except UCCA. SJTU-NICT \cite{li2019sjtu} and SUDA–Alibaba \cite{zhang2019suda} adopt a graph-based UCCA parser \cite{jiang2019hlt}.  SJTU–NICT \cite{li2019sjtu}, SUDA–Alibaba \cite{zhang2019suda}, and Amazon \cite{cao2019amazon} remove remote edges before the models. Then the remote edges are added in the post-processing step and a separate classifier is used for prediction of remote edges.  Hitachi \cite{koreeda2019hitachi} is a unified encoder-to-biaffine network with a shared architecture, and JBNU \cite{na2019jbnu} is a unified parsing model that is based on a biaffine attention with a BiLSTM encoder and a biaffine-attention decoder. The model uses a bilexical format for UCCA in training by converting UCCA into this format (bilexical format) as a preprocessing step. 

Composition-based approach \cite{che2019hit,donatelli2019saarland,oepen2019erg} has also been utilized for UCCA representation. Saarland \cite{donatelli2019saarland} propose a parser that learns to map sentences to  Apply-Modify (AM) dependency trees instead of learning to map directly from sentences to graphs. The other model ERG \cite{oepen2019erg} applies English Resource Semantics (ERS) parsing.

In MRP 2020 \cite{oepen2020mrp}, the participant models are mainly based on encoder-decoders with slight variations in the decoder components. Hitachi \cite{dou2020hit} is a  transformer-based encoder-decoder architecture. It utilizes a Plain Graph Notation (PGN) within the decoder. The proposed parser predicts a PGN based sequence by leveraging transformers and biaffine notations. \'UFAL \cite{samuel2020ufal} is a novel permutation-invariant approach (PERIN) proposed for sentence-graph semantic parsing. The authors compute contextualized embeddings with  XLM-RoBERTa (XLM-R) in the encoder and use an attention head to predict the edges and edge labels. HIT-SCIR \cite{dou2020hit} and JBNU \cite{na2020jbnu} adopt iterative inference framework proposed by \cite{cai2020amr}. The iterative inference framework is also based on encoder-decoder architecture. While HIT-SCIR \cite{dou2020hit} uses the iterative inference with little modification, JBNU \cite{na2020jbnu} extends the framework by adding a shared state. Finaly, JUJI-KU \cite{arviv2020huji} uses TUPA parser at MRP \cite{hershcovich2019tupa} and HIT-SCIR \cite{che2019hit}. 
\section{Shared Tasks}\label{sec:shared}

The goal of the shared tasks is to benefit semantic parsing in resource-scarce languages or frameworks by using data in resource-rich languages or frameworks. We can classify the shared tasks that use UCCA representation into two categories: Cross-Lingual and Cross-Framework. Cross-lingual parsing \cite{zhang2017selective, zhang2018cross} is the task of parsing where we train a parser on a source language and transfer it to a target language. Cross-Framework \cite{dou2020hit, hershcovich2018multitask, samuel2020ufal} is the task of parsing a text for more than one meaning representation framework. While cross-lingual learning aims to improve the results for resource-scarce languages, the cross-framework learning aims to reduce the number of framework-specific methods. 

SemEval 2019 \cite{hershcovich2019semeval} and MRP 2019 \cite{oepen2019mrp} are built on cross-lingual and cross-framework tasks respectively, and MRP 2020  \cite{oepen2020mrp} involves tracks on cross-lingual and cross-framework. The main challenge in cross-framework semantic parsing task is that diverse frameworks differ in mapping methods between surface strings and graph nodes, which incurs the incompatibility among framework-specific parsers. The frameworks deviate from each other with their node labels and properties. MRP 2019 and MRP 2020 use directed graphs to unify the five different semantic representation frameworks to overcome this problem. MRP format is used along with a common JSON-based interchange format that allows an arbitrary graph structure whose nodes may or may not be anchored to the input text's spans. 


\section{Self-Attentive UCCA-based Parsing}\label{sec:model}

We propose a neural model for semantic parsing that tackles the task as a constituency parsing problem. Our model is inspired by non-projective dependency parsing by \cite{nilsson2005pseudo}, which was also used on semantic parsing approaches \cite{jiang2019hlt,zhang2019suda}.

\subsection{Constituency Parsing}

We adopt the constituency parsing model based on self-attention mechanism proposed by \cite{kitaev2018multilingual}. It follows the chart-based constituency parsing where the constituency tree $T$ of an input sentence $s= \{w_1, \cdots, w_n\}$ is represented as a set of labeled spans:
\begin{equation}
    T = \{(i_t, j_t, l_t): t = 1, \cdots, |T|\}
\end{equation}
where $(i_t,j_t,l_t)$ is produced by the encoder where $i_t$ and $j_t$ refer to the beginning and ending positions of the $t^{th}$ span  with a label $l_t \in L$. 

We assign a score $s(T)$ to each tree, which is decomposed as follows:
\begin{equation}
    s(T) = \sum_{(i,j,l) \in T} s(i,j,l)
\end{equation}

\begin{figure}[th!]
\centering 
\resizebox*{8 cm}{!}{\includegraphics{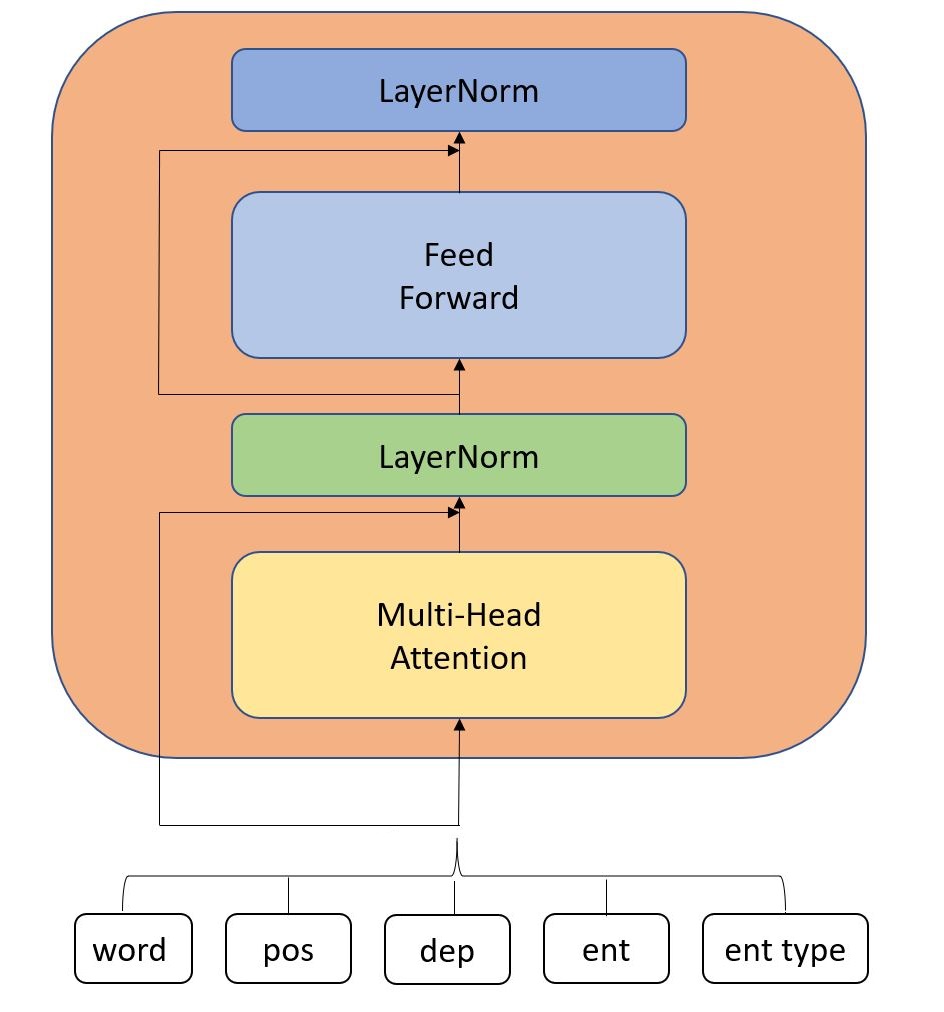}}
\caption{An overview of the self-attention encoder, which has 8 identical layers}\label{fig:encoder}
\end{figure}

The encoder in the model involves self-attention layers that learn a context vector $y_t$ for each position $t$ for each word vector $x_t$. The overview of one layer of encoder architecture is given in Figure \ref{fig:encoder}. Our encoder involves $8$ layers.

Each word $w_t$ is mapped into a dense vector $x_t$:
\begin{equation}
    x_t = e_{w_t} \oplus e_{p_t} \oplus e_{d_t}  \oplus e_{e_t} \oplus e_{e_ob_t}
\end{equation}
which is a concatenation of the word embedding $e_{w_t}$, PoS tag embedding $e_{p_t}$, dependency label embedding $e_{d_t}$, entity type embedding $e_{e_t}$ , and entity iob type embedding $e_{e_ob_t}$.  

\begin{figure}[h!]
\centering 
\resizebox*{8 cm}{!}{\includegraphics{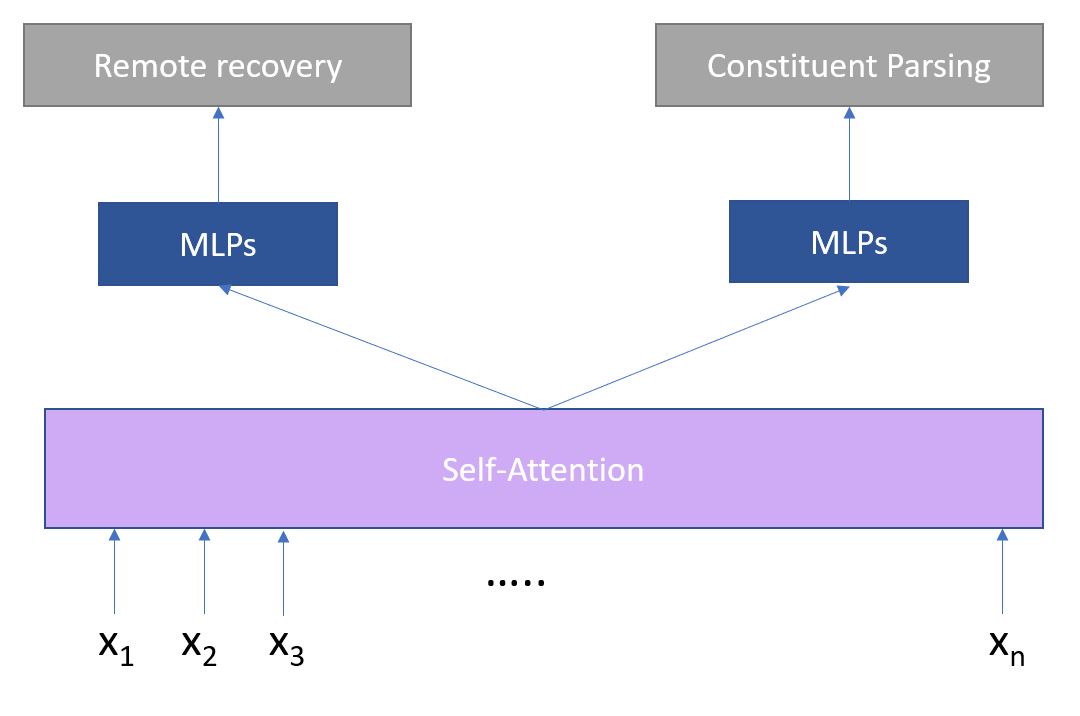}}
\caption{Remote Edge Recovery}\label{fig:remote}
\end{figure}

An MLP classifier consisting of two fully-connected layers with ReLU nonlinearity assigns labeling scores s(i,j,l) to each span. We utilize a remote edge recovery model that also shares the same encoder to recover remote edges in trees \cite{jiang2019hlt} as given in Figure~\ref{fig:remote}. Therefore, the model incorporates two independent MLPs to predict remote edges and candidate parent nodes using the same encoder.

The parsing loss is the summation of the cross-entropy losses introduced by the remote edges and non-terminal node pairs:

\begin{equation}
    L = L_{remote} + L_{non-terminal}
\end{equation}

\begin{figure}[h!]
\centering 
\resizebox*{10 cm}{!}{\includegraphics{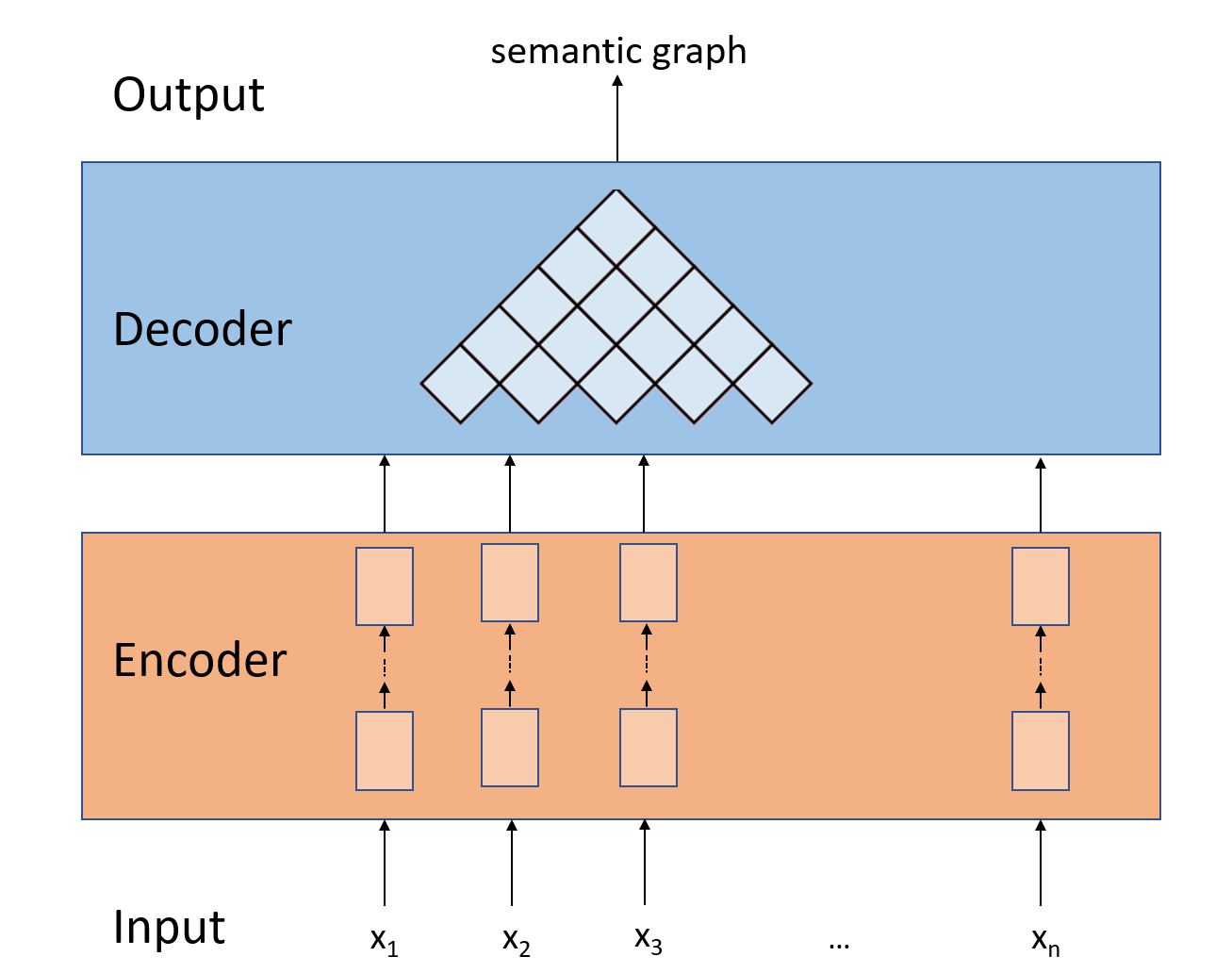}}
\caption{The general architecture of the proposed model}\label{fig:architecture}
\end{figure}

As for the inference, CYK algorithm is used to generate a globally optimized tree $ \hat{T}$ for each sentence that is used as a decoder in the model (Figure~\ref{fig:architecture}):
\begin{equation}
    \hat{T} = \argmax{T} s(T)
\end{equation}

\section{Experiments \& Results}\label{sec:result}

We did experiments in two folds: single-lingual and cross-lingual.

\subsection{Datasets}

We used SemEval 2019 shared task dataset  \cite{hershcovich2019semeval} in this study because it is especially built for cross-lingual parsing. The details of the datasets are given in Table \ref{tab:dataset}.  It involves datasets in English, German, and French.  We performed two single-lingual experiments similar to \cite{hershcovich2019semeval} for English: 1. In-domain setting using English-Wiki corpus for training and validation set of English-Wiki for testing purposes, 2. Out-of-domain setting using English-Wiki corpus for training and validation set as English-20K for testing purposes. We only performed in-domain experiments for German and French, since only one dataset is available for the two languages. 

Because of insufficient training data available for French in SemEval dataset, we performed cross-lingual experiments by merging training datasets of three languages to train the model in a cross-lingual setting. In this way, we expect the model to utilize all training data as if it is a single language, which helps sharing the languages' parameters using the similarities in different languages while learning the language differences concurrently.

\begin{table}[ht!]\centering
    \caption{Number of sentences in each dataset in Semeval 2019 Dataset \cite{hershcovich2019semeval}}
    \centering
    \begin{tabular}{lcccc}
\toprule
 & English-Wiki & English-20K & German-20K & French-20K \\ \midrule
Train                  & 4,113        & 0           & 5,211      & 15         \\ 
Validation             & 514          & 0           & 651        & 238        \\ 
Test                   & 515          & 492         & 652        & 239        \\ \bottomrule
    \end{tabular}

    \label{tab:dataset}
\end{table}

\subsection{Hyperparameters and Implementation Details}

For the encoder, we used a self-attention layer with the same parameter values used in \cite{vaswani2017attention}. The word embedding dimensionality is $100$ with an embedding dimensionality of $50$, for PoS tags, $50$ for dependency tags, $25$ for entity types, and $25$ for the entity iob types. We used Adam \cite{kingma2014adam} optimizer and early stopping because of the variations in the size of the training sets.

All syntactic embeddings (word, PoS tags, dependency tags, entity types, and entity iob types) are randomly initialized for single-lingual experiments. In addition to the syntactic embeddings, we used pre-trained fasttext \cite{bojanowski2016enriching} character n-gram based word embeddings. Additionally, we used BERT embeddings as contextualized embeddings to incorporate contextual information. For the cross-lingual models, we did experiments with and without contextual embeddings in addition to the syntactic embeddings in the model.

\subsection{Evaluation and Results}

 We followed the official evaluation metrics provided by \cite{hershcovich2019semeval} for SemEval 2019. The evaluation method measures a matching score between each output graph $G_o = (V_o, E_o,l_o)$ and its corresponding gold graph $G_g = (V_g, E_g,l_g)$  over the same sequence of nodes. Labeled precision and recall metrics are calculated by dividing the number of matching edges in $G_o$ and $G_g$ with their corresponding labels to $|E_o|$ and $|E_g|$ respectively.
 
 $F_1$ is the harmonic mean of precision and recall:
 \begin{equation}
    F_1 = 2 \cdot \frac{Precision \cdot Recall}{Precision \times Recall}
 \end{equation}
 
Unlabeled precision, recall, and $F_1$ are calculated analogously, but without requiring label matching for the edges. We evaluate both primary and remote edges separately in all experiments.
 
The results of both single-lingual and cross-lingual experiments on  SemEval 2019 \cite{hershcovich2019semeval} datasets for English, French and German are given in Table \ref{tab:experiments}.

\begin{table}[ht!]\centering
 \caption{Single-Lingual and Cross-Lingual Experimental Results on Semeval 2019 dataset }
\begin{tabular}{lcccccc}
\toprule
\multicolumn{1}{c}{}             & \multicolumn{3}{c}{Single-Lingual Exp.}               & \multicolumn{3}{c}{Cross-Lingual Exp.}   \\    \\ \toprule
         & Prim. & Rem.   & Avg   & Prim.    & Rem.   & Avg   \\ \midrule

   & \multicolumn{6}{c}{English-Wiki}    \\ \midrule
syntactic emb.  & 74.5  & 2.1   &     73.04  & 74.8    & 44.7    &    74.19     \\
syntactic emb. $\oplus$ fasttext   & 77.9      & 53.0 &   77.4  & - &- & -   \\ 
syntactic emb. $\oplus$ bert  & 78.3   & 52.8   &77.79  & 79.6   & 48.5   &  78.97 \\ 
syntactic emb. $\oplus$ fasttext $\oplus$ bert   & 80.2  & 55.4  & \textbf{79.7}   & -   & -   & -   \\ \bottomrule
 & \multicolumn{6}{c}{English-20K}  \\ \toprule
syntactic emb.  & 71.0  & 7.9&   68.87 & 72.7  & 23.6   &   71.04   \\ 
syntactic emb. $\oplus$ fasttext   & 73.8 & 25.0    &    72.15        & - & -    & -  \\ 
syntactic emb. $\oplus$ bert    & 75.45   & 28.6    & 73.87  & 75.9 & 29.4   & 74.33  \\ 
syntactic emb. $\oplus$ fasttext $\oplus$ bert  & 76.2  & 29.3  &   \textbf{74.62}  & -  & -    & -  \\ \bottomrule
  & \multicolumn{6}{c}{German-20K}   \\ \toprule
syntactic emb.  & 77.3   & 31.5  &   76.09   & 80.4  & 49.3  &  79.58  \\ 
syntactic emb. $\oplus$ fasttext & 83.6   & 60.2  &     82.98   & -   & -   & -   \\ 
syntactic emb. $\oplus$ bert     & 85.1    & 63.7    & 84.54 & 86.2 & 53.6    & 85.34  \\ 
syntactic emb. $\oplus$ fasttext $\oplus$ bert   & 86.7  & 65.1   & \textbf{86.13 }  & -    & -   & -  \\ \bottomrule
 & \multicolumn{6}{c}{French-20K} \\ \toprule
syntactic emb. & 43.1  & 0  &   41.67  & 65.4 & 15.3 &    63.74    \\ 
syntactic emb. $\oplus$ fasttext & 43.2   & 0  &41.77   & -   & -    & - \\
syntactic emb. $\oplus$ bert   & 44.5     & 0    & 43.02   & 68.7   & 45.5    &  \textbf{ 67.93}\\ 
syntactic emb. $\oplus$ fasttext $\oplus$ bert  & 46.2  & 0    & 44.67  & -  & -    & -  \\ \bottomrule
\end{tabular}
 \label{tab:experiments}
\end{table}

For the single-lingual setting, using fasttext embeddings along with BERT contextualized embeddings in addition to the syntactic embeddings outperforms other settings on all languages. For the cross-lingual setting, the results of languages except French have dropped slightly. However, French results improved significantly. The cross-lingual setting helps to predict remote edges in French, while it is not sufficient to predict remote edges due to the inadequate amount of training data for French in single-lingual setting. We did not run experiments by using pre-trained fasttext embeddings because they are independently trained for different languages and not available as multilingual embeddings. Using BERT improves F1 scores by around 4\% for English, 5\% for German, and 4\% for French in cross-lingual setting. 

 Comparative results of our model with other participants at Semeval 2019 \cite{hershcovich2019semeval}\footnote{We give the official results given in \cite{hershcovich2019semeval}} are given in Table \ref{tab:all-result}. The results show that our model achieves state-of-the-art performance among the other parsers in English and German. \cite{jiang2019hlt} outperforms other models in French. However, our results are still competitive to that of \cite{jiang2019hlt} on unlabeled edges. 

 \begin{table}[htbp]
     \centering
          \caption{Comparative F-1 results of our model with other participants of UCCA framework at Semeval 2019}
     \begin{tabular}{lcccccc}
\toprule
      & \multicolumn{6}{c}{English-Wiki} \\\midrule
      & \multicolumn{3}{c}{Labeled} & \multicolumn{3}{c}{Unlabeled}\\\midrule
       & All & Prim. & Rem. & All & Prim. & Rem.\\\midrule
      Tupa    & 72.8  & 73.3 & 47.2 & 85.0& 85.8& 48.4\\
      HLT@SUDA & 77.4 & 77.9 & 52.2 & 87.2 &87.9 & 52.5 \\
      Davis & 72.2  &   73.0    & 0 & 85.5 &  86.4 &  0  \\
      CUNY-PekingU & 71.8 & 72.3 & 49.5 & 84.5 & 85.2 & 50.1  \\
       DANGNT@UIT.VNU-HCM &   70.0 & 70.7 & 0 & 81.7 & 82.6 & 0  \\
        GCN-Sem & 65.7 & 66.4 & 0 & 80.9 & 81.8 & 0 \\\toprule
     Self-Attentive UCCA Parser  & \textbf{79.7}& 80.2 & 55.4 & \textbf{89.6} &  90.3& 55.3 \\\bottomrule
      & \multicolumn{6}{c}{English-20K} \\\toprule
          HLT@SUDA &72.7&73.6&31.2&85.2&86.4&32.1\\
    Tupa&67.2&68.2&23.7&82.2&83.5&24.3\\
    CUNY-PekingU&66.9&67.9&27.9&82.3&83.6&29.0\\
    GCN-Sem&62.6&63.7&0&80.0&81.4&0\\\midrule
      Self-Attentive UCCA Parser   & \textbf{74.62} & 76.2 & 29.3&  \textbf{87.69} & 89.7& 30.1\\\bottomrule
      & \multicolumn{6}{c}{German-20K} \\\toprule
          HLT@SUDA&84.9&85.4&64.1&92.8&93.4&64.7\\
    Tupa&79.1&79.6&59.9&90.3&91.0&60.5\\
    TüPa&78.1&78.8&40.8&89.4&90.3&41.2\\
    XLangMo&78.0&78.4&61.1&89.4&90.1&61.4\\
     MaskParse@Deski\~n & 74.2  & 74.8 & 47.3 & 87.1 & 88.0 & 47.6 \\\midrule
      Self-Attentive UCCA Parser   &\textbf{86.13}  & 86.7 & 65.1 & \textbf{94.1} & 94.4 & 64.5 \\\bottomrule
      & \multicolumn{6}{c}{French-20K} \\\midrule
          HLT@SUDA&\textbf{75.2}&76.0&43.3&\textbf{86.0}&87.0&45.1\\
    XLangMo&65.6&66.6&13.3&81.5&82.8&14.1\\
    MaskParse@Deski\~n &65.4&66.6&24.3&80.9&82.5&25.8\\
    Tupa&48.7&49.6&2.4&74.0&75.3&3.2\\
    TüPa&45.6&46.4&0&73.4&74.6&0\\
     MaskParse@Deski\~n & 65.4 & 66.6 & 24.3 & 80.9 & 82.5 & 25.8 \\\midrule
      Self-Attentive UCCA Parser   & 67.93 &68.7  &45.5 & 84.8  & 85.5 & 54.6\\\bottomrule
     \end{tabular}

     \label{tab:all-result}
 \end{table}

\textbf{Zero-shot and Few-shot Cross-Lingual Model:} Zero-shot learning \cite{wang2019cross, tran2019zero} and few-shot learning \cite{lauscher2020zero} have shown outstanding success recently in various NLP tasks such as dependency parsing and text classification. Zero-shot cross-lingual model is used when there is no or few annotated examples are available in the target language. In contrast, few-shot cross-lingual model is used when a small amount of training data is available during training. We performed both few-shot and zero-shot learning on the French dataset as part of the cross-lingual experiments due to its insufficient size.

In zero-shot setting, we ran cross-lingual experiments without using the French dataset, whereas we included the dataset in few-shot learning setting. The results are given in Table \ref{tab:result-data}.
The results show that even a small amount of data improves the results substantially in few-shot learning compared to zero-shot learning.

\begin{table}[ht!]\centering
\centering
     \caption{Effect of French dataset on cross-lingual model}
\begin{tabular}{lcccccc}
\toprule
\multicolumn{1}{c}{} & \multicolumn{3}{c}{Labeled} & \multicolumn{3}{c}{Unlabeled} \\\midrule
\multicolumn{1}{c}{} & Primary   & Remote  & Avg    & Primary   & Remote   & Avg     \\ \midrule
single-lingual         &  46.2         &    0     &     44.67   &    68.9       &  0        &   66.62      \\ 
zero-shot              & 57.18     & 16.2    & 56.42  & 78.6      & 16.2     & 76.53   \\ 
few-shot               & \textbf{68.7}      & \textbf{45.5}    & \textbf{67.93}  & \textbf{85.8}      & \textbf{54.6}     & \textbf{84.48}   \\ \bottomrule
\end{tabular}

     \label{tab:result-data}
\end{table}

     \subsection{Error Analysis}

We characterize errors made by our parser by performing further experiments to analyze the impact of structural and linguistic features of sentences on the parser's accuracy.

\textbf{Sentence Length:} The results according to different lengths of sentences are given in Table \ref{tab:result-length}. The results show that the longer the sentences are, the lower the F-1 scores are for the remote edges except for the English-Wiki dataset. However, remote edge scores on the English-Wiki dataset are higher in longer sentences. 

The frequency of a remote edge is $1$ or $0$ in each sentence in the dataset. Therefore, the strength of the model on primary edges is more important compared to remote edges.

\begin{table}[ht!]\centering
\footnotesize
\centering
     \caption{Result by Length}
\begin{tabular}{ccccccccccccc}
\toprule
                       & \multicolumn{3}{c}{English-Wiki}                                                 & \multicolumn{3}{c}{English-20K}                                                  & \multicolumn{3}{c}{German-20K}                                                   & \multicolumn{3}{c}{French-20K}                                                   \\ \midrule
\multicolumn{1}{c}{Sent. Len.} & \multicolumn{1}{c}{Prim.} & \multicolumn{1}{c}{Rem.} & \multicolumn{1}{c}{All} & \multicolumn{1}{c}{Prim.} & \multicolumn{1}{c}{Rem.} & \multicolumn{1}{c}{All} & \multicolumn{1}{c}{Prim.} & \multicolumn{1}{c}{Rem.} & \multicolumn{1}{c}{All} & \multicolumn{1}{c}{Prim.} & \multicolumn{1}{c}{Rem.} & \multicolumn{1}{c}{All} \\ \midrule
\textgreater{}=10      & 80.8                       & 50.8                      & 79.76                    & 73.2                       & 30.1                      & 72.33                    & 87.2                       & 66.7                      & 86.61                    & 68.2                       & 45.4                      & 67.42                    \\ 
\textgreater{}=20      & 80.4                       & 50.9                      & 79.42                    & 76.9                       & 29.0                      & 75.89                    & 84.5                       & 60.3                      & 83.75                    & 67.5                       & 45.3                      & 66.72                    \\ 
\textgreater{}=30      & 79.8                       & 63.0                      & 79.18                    & 75.8                       & 26.6                      & 74.74                    & 83.4                       & 59.2                      & 82.67                    & 69.1                       & 42.7                      & 68.17                    \\ 
\textgreater{}=40      & 82.4                       & 61.0                      & 81.68                    & 74.5                       & 26.4                      & 73.58                    & 83.0                       & 48.5                      & 82.01                    & 66.5                       & 41.4                      & 65.58                    \\ 
\textgreater{}=50      & 78.5                       & 61.5                      & 77.93                     & 74.2                       & 0                         & 72.55                    & 82.9                       & 49.2                      & 81.84                    & 66.8                       & 39.8                      & 65.83                    \\ \bottomrule
\end{tabular}

     \label{tab:result-length}
\end{table}

\textbf{Semantic Categories:} We analyze each semantic category's results to further assess the model's performance according to the categories. The results are given in Table \ref{tab:result-type}. The frequencies of Adverbial (A), Function (F), Ground (G), Linker (L), Connector (C), and State (S) categories are comparably lower than other semantic categories in the dataset\footnote{The dataset details are given in \cite{hershcovich2019semeval}.}.   

While the model struggles with predicting categories with low frequency, more frequent categories are learned more accurately.

\begin{table}[!ht]
\footnotesize
\centering
     \caption{F-1 measure of predicting Primary edges and their labels}
\begin{tabular}{ccccccccccccc}
\toprule
dataset      & A    & C    & D    & E    & F    & G    & H    & L    & N    & P    & R    & S    \\ \midrule
English-Wiki & 0.77 & 0.83 & 0.67 & 0.80 & 0.73 & 0.65 & 0.75 & 0.56 & 0.87 & 0.68 & 0.86 & 0.24 \\
English-20K  & 0.67 & 0.81 & 0.54 & 0.78 & 0.71 & 0.25 & 0.60 & 0.72 & 0.82 & 0.73 & 0.86 & 0.23 \\
German-20K   & 0.81 & 0.90 & 0.77 & 0.87 & 0.88 & 0.71 & 0.79 & 0.88 & 0.30 & 0.79 & 0.92 & 0.27 \\ 
French-20K   & 0.58 & 0.78 & 0.32 & 0.71 & 0.41 & 0.46 & 0.49 & 0.59 & 0.75 & 0.68 & 0.83 & 0.24 \\ \bottomrule
\end{tabular}

     \label{tab:result-type}
\end{table}

\section{Conclusion and Future Work}\label{sec:conc}

In this paper, we propose a constituency parsing model for learning UCCA semantic representation, which is based on a self-attention mechanism where various types of syntactic and semantic embeddings are incorporated in training. We obtain state-of-the-art parsing results with a self-attention encoder on English and German in a single-lingual model.

Moreover, we propose a cross-lingual model to compare the performance with other models proposed for the UCCA framework. The results show that the cross-lingual model performs better on low-resource languages. The results on the French language are remarkably better in the cross-lingual model. Our models suggest that further research in cross-lingual learning has the potential to lead to improvements on creating a dataset for no resource languages.

\bibliographystyle{unsrt}  
\bibliography{references}

\end{document}